%% file: ICRA-40-Galileo.tex
\documentclass[letterpaper, 10 pt, conference]{ieeeconf}
\IEEEoverridecommandlockouts
\usepackage{cite}
\usepackage{amsmath,amssymb,amsfonts}
\usepackage{mathtools}
\usepackage{graphicx}
\usepackage{float}
\usepackage{mathrsfs}
\usepackage{hyperref}
\usepackage{subcaption}
\title{\bf{Galileo: A Pseudospectral Collocation Framework for Legged Robots}}

\author{Ethan Chandler$^{*}$, Akshay Jaitly$^{*}$, Mahdi Agheli$^{*}$%
\thanks{$^{*}$Authors are with the Department of Robotics Engineering at the Worcester Polytechnic Institute, Worcester, MA, 01609, USA. Corresponding Author: {\tt\small mmaghelih@wpi.edu}}%
}

\begin{document}

\maketitle

\input{Abstract}

\input{Introduction}


\input{Galileo}

\input{Galileo-Legged}

\input{Results}

\input{Conclusions}

\bibliographystyle{IEEEtran} 
\bibliography{refs} 

\end{document}

%% file: Abstract.tex
\begin{abstract}\label{abstract}
Dynamic maneuvers for legged robots present a difficult challenge due to the complex dynamics and contact constraints. This paper introduces a versatile trajectory optimization framework for continuous-time multi-phase problems. We introduce a new transcription scheme that enables pseudospectral collocation to optimize directly on Lie Groups, such as SE(3) and quaternions without special normalization constraints. The key insight is the change of variables---we choose to optimize over the history of the tangent vectors rather than the states themselves. Our approach uses a modified Legendre-Gauss-Radau (LGR) method to produce dynamic motions for various legged robots. We implement our approach as a Model Predictive Controller (MPC) and track the MPC output using a Quadratic Program (QP) based whole-body controller. Results on the Go1 Unitree and WPI’s HURON humanoid confirm the feasibility of the planned trajectories.

\end{abstract}

%% file: Introduction.tex
\section{Introduction}\label{introduction}

Trajectory synthesis presents a fundamental challenge in robotics. The main goal is to determine control inputs that guide a robot from an initial to a desired state while navigating a constraint-rich state space \cite{Tedrake2023}. This is particularly difficult for legged robots, due to their complex multi-phase dynamics. These restrictions motivate the need to automatically generate valid trajectories for dynamic motions.

Collocation methods solve the trajectory optimization problem via function approximation \cite{Kelly2017}. In this paper, we will focus on pseudospectral collocation \cite{Gong2007, Garg2009, Garg2010}, which approximates the optimal state and control trajectory with orthogonal polynomials, such as Lagrange or Chebyshev polynomials. Pseudospectral collocation achieves convergence by optimizing the state and the controls at specially defined collocation points---typically, the polynomial roots. Unfortunately, the pseudospectral collocation methods presented in these works do not immediately lend themselves to interpolation over manifolds.

We introduce a simple yet holistic solution to this problem by optimizing over the tangent space. Optimizing over the tangent space rather than the state manifold is not new---stemming primarily from approaches seeking to parameterize the $SO(3)$ rotation matrix such as \cite{Nguyen2022, Ding2021, Shen2022}. However, these approaches are only performed using forward Euler transcription, which is known to have substantial drift over practical time horizons \cite{Bordalba2023}.

\begin{figure}[t]
\centering
{\includegraphics[width=0.65\columnwidth]{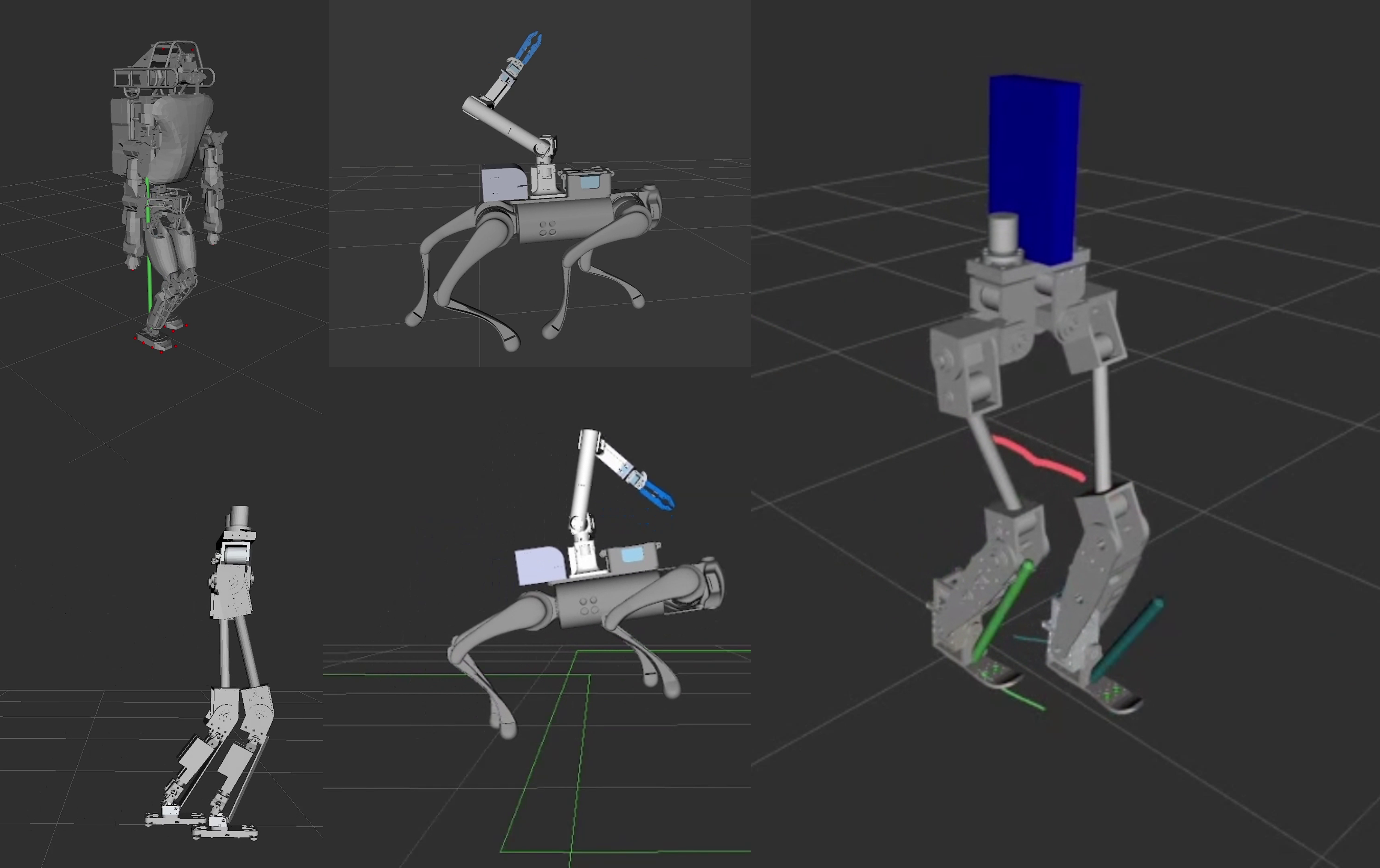}}
\caption{Galileo MPC solving legged robot motions at 50Hz, simulated in Gazebo. Video results shown in \cite{galileo}.}
\label{fig:galileo-results}
\end{figure}

Our contributions include: 1) Galileo \cite{galileo}: A lightweight, extensible, and open-source optimization library for switched systems, based on LGR pseudospectral collocation, 2) A new transcription method to enable pseudospectral collocation over differentiable manifolds, without the need for additional normalization constraints, and 3) A Galileo MPC interface for legged robots, including simulation examples for the Unitree Go1 quadruped, Atlas, and WPI's HURON.

%% file: Galileo.tex
\section{Galileo}\label{galileo}

The purpose of Galileo is to transcribe multi-phase trajectory optimization problems into ones solvable by nonlinear program (NLP) solvers, such as IPOPT \cite{ipopt}. 

\subsection{Multi-phase Optimization Problem Formulation} \label{multi_phase}

Given a sequence of phases, $[1, ..., P]$, where $p$ represents a distinct segment of the trajectory with its own dynamics and constraints, we seek to solve the following class of trajectory optimization problems:

\begin{flalign}
\underset{\mathbf{x}, \mathbf{u}}{\text{minimize}} \hspace{6pt} \Phi(\mathbf{x}(t_{f})) + \sum_{p=1}^{P}\Biggl[ \int_{t_{0,p}}^{t_{f,p}}\mathcal{L}_{p}(\mathbf{x}(t), \mathbf{u}(t))dt \Biggr] \label{eq:cost}&&
\end{flalign}
\vspace{-15pt}
\begin{flalign}
\text{subject to} \hspace{6pt} & \dot{\mathbf{x}}(t) = \mathbf{f}_{p}(\mathbf{x}(t), \mathbf{u}(t)) &\forall t \in [t_{0,p}, t_{f,p}] \label{eq:dynamics} &&\hspace{0pt}\\
& \mathbf{c}_{p}(\mathbf{x}(t), \mathbf{u}(t)) \le \mathbf{0} &\forall t \in [t_{0,p}, t_{f,p}] \label{eq:ineq} &&\\
& \mathbf{b}(\mathbf{x}(t_{0}),\mathbf{x}(t_{f})) = \mathbf{0} & \label{eq:bound} &&\\
& \mathbf{x}(t_{0,p+1}) = \mathbf{S}_{p}(\mathbf{x}(t_{f,p}))\hspace{-11pt} & \label{eq:jump_map} &&\\
\nonumber& \forall p \in [1,...,P] & &&
\end{flalign}

This is known as multi-phase Bolza form \cite{Tedrake2023}, where $t_{0,p}$ and $t_{f,p}$ represent the boundary times of phase $p$. Here $\mathcal{L}_{p}$ and $\Phi$ in \eqref{eq:cost} are the incremental cost for each phase and the terminal cost, respectively. \eqref{eq:dynamics} and \eqref{eq:ineq} enforce the dynamics and inequality constraints associated with phase $p$, and \eqref{eq:bound} constrains the initial and final states in the trajectory. \eqref{eq:jump_map} is the jump map, which is a constraint enacted on the transitions between consecutive phases.

Overall, the optimization seeks a trajectory $\mathbf{x}$ and control inputs $\mathbf{u}$ that minimize the cost while satisfying the active constraints in each phase. Efficiently solving this class of trajectory optimization problems is the focus of this paper.

\begin{figure}[t]
\centering
{\includegraphics[width=0.85\columnwidth]{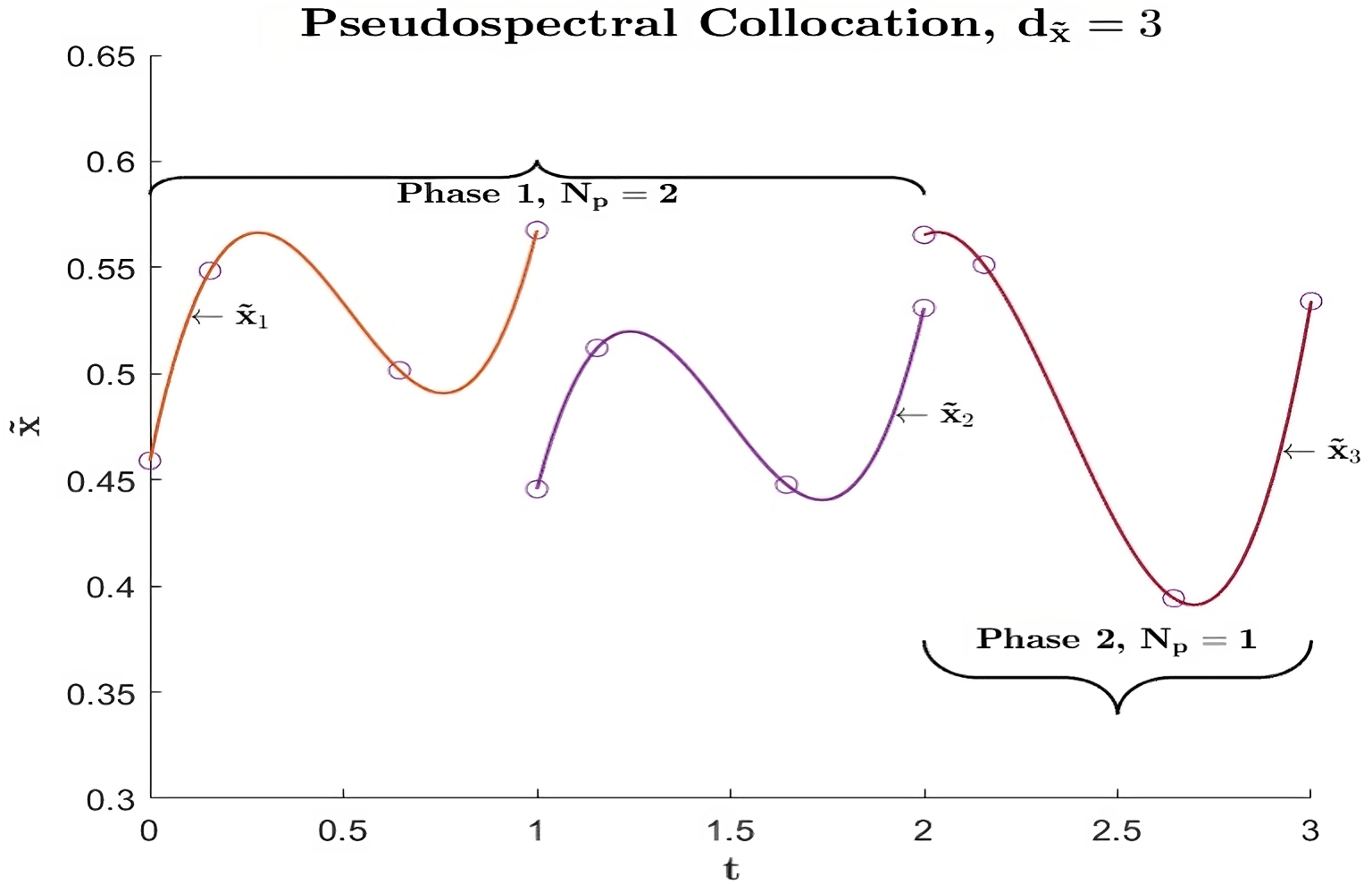}}
\caption{Visualization of the polynomial approximation used in pseudospectral collocation.}
\label{fig:pseudospectral-collocation}
\end{figure}

\subsection{Optimizing with Deviations} \label{optimizing}
Practical applications in robotics often require subsets of the decision variables to lie on differentiable manifolds. Quaternions, for instance, are required to stay on a hypersphere in $\mathbb{R}^4$, which is difficult to enforce without employing explicit normalization constraints at each iteration.

Our method directly optimizes over these differentiable manifolds---without the use of additional equality constraints---through a change in decision variables. Instead of explicitly including manifold states (orientations, for instance) as our decision variables, we use the state deviants, which are tangents on the state manifold. These state deviants will be referred to with $\mathbf{\tilde{x}}$. Thus, our decision variables are $\tilde{\mathbf{x}}$ and $\mathbf{u}$.

Modifying the Gaussian quadrature used in \cite{Kelly2017}, we transcribe the continuous-time objective \eqref{eq:cost} with:
\begin{equation}
    J_{p} =\sum^{N_p}_{k=0}\Biggl[\sum^{d_{\tilde{x}}}_{j=0}\Biggl[B_{\tilde{x}, j}\mathcal{L}_{p}(\mathbf{F}_{\text{exp}}(\mathbf{x}_{0}, \mathbf{\tilde{\mathbf{x}}}_{k,j}), \mathbf{u}(\tau_{\tilde{x},j}))\Biggr]\Biggr]
\end{equation}

where $\mathbf{F}_{\text{exp}}$ is the differentiable manifold's exponential map, and $\mathbf{\tilde{\mathbf{x}}}_{k,j}$ refers to the state deviant at the $k$-th knot segment and $j$-th collocation point (Fig. \ref{fig:pseudospectral-collocation}), determined by the Legendre-Gauss-Radau (LGR) points \cite{Garg2009}. Here, $B_{\tilde{x}, j}$ is the $j$-th element of the quadrature coefficients for the Lagrange polynomial of degree $d_{\tilde{x}}$. $\mathbf{u}(\tau_{\tilde{x},j})$ refers to the control input at the state deviant collocation time $\tau_{\tilde{x},j}$ (mapped to the range $[0, 1]$) on the control input's Lagrange polynomial. For numerical accuracy, this control input value is determined using Barycentric Lagrange interpolation.

In order to ensure that the trajectory satisfies the system dynamics, we transcribe \eqref{eq:dynamics} with the collocation constraint
\begin{equation}
    h \cdot \mathbf{f}_{p}(\mathbf{F}_{\text{exp}}(\mathbf{x}_{0}, \tilde{\mathbf{x}}_{k,j}), \mathbf{u}(\tau_{\tilde{x},j})) - \mathbf{D}_{\tilde{x}}\tilde{\mathbf{x}}_{k} = 0
    \label{galileo_dynamics}
\end{equation}

where $\mathbf{D}_{\tilde{x}}$ is the differentiation matrix corresponding to the Lagrange polynomial of degree $d_{\tilde{x}}$ and $h$ is the timestep. Transcription of the inequality \eqref{eq:ineq} and boundary constraints \eqref{eq:bound} follows similar logic.

The number of decision variables for our transcription is $\sum^{P}_{p=0}\left(N_{p}(n_{\tilde{x}}d_{\tilde{x}} + n_{u}d_{u}) + (N_{p}+1) (n_{\tilde{x}}+n_{u})\right)$ and the number of equality constraints is $\sum^{P}_{p=0}\left(n_{x} + N_{p}(n_{x}d_{x}+n_{\tilde{x}}+n_{u})\right)$. This is superior to normalization-based methods which add an additional $\sum^{P}_{p=0}\left(N_{p}n_{x}d_{x}\right)$ equality constraints to ensure that the state lies on the manifold.

%% file: Galileo-Legged.tex

\section{Galileo for Legged Robot Control}\label{galileo_dynamics}
To use our transcription scheme for legged robot control, we define our state with $\mathbf{x} = \left[\mathbf{k}_{\text{com}}^{\intercal}, \mathbf{q}_b^{\intercal}, \mathbf{q}_j^{\intercal}\right]^{\intercal} \label{eq:state_def}$
where $\mathbf{k}_{\text{com}}$ is the centroidal momentum about the center of mass, $\mathbf{q}_b $ is the base position and orientation w.r.t the fixed inertial frame, and $\mathbf{q}_j$ is the joint configuration. The base orientation is represented as a quaternion. For the remainder of this paper, we shall refer to the quaternion component of $\mathbf{q}_b$ with $\mathbf{q}_{bo}$.


For the control input, we define $\mathbf{u} = \left[\mathbf{f}_{\text{wrench,i}}^{\intercal},  \mathbf{v}_{j}^{\intercal}\right]^{\intercal}$ 
where $\mathbf{f}_{\text{wrench,i}}$ is the stacked contact force $\mathbf{f}_{e_{i}}$ and contact torque $\boldsymbol{\tau}_{e_{i}}$ for end effectors $i$ with contact patches, or simply the contact force at the $i$-th end effector if it is a point contact. $\mathbf{v}_{j}$ is the joint velocity variable.

We use Centroidal Momentum Dynamics for the legged robot problem due to the physical accuracy compared to simpler models like Single Rigid Body (SRB), and the reduced computation cost compared to the whole-body dynamics.

\subsection{Quaternion Parameterization for Legged Robot Pose}

For the legged robot problem, we use position vectors and orientation quaternions to describe the pose. Thus, $\mathbf{q}_{bo}$ is restricted to lie on $\mathcal{H}$, which is a hypersphere in $\mathbb{R}^{4}$. 

We define the state deviants with $    \mathbf{\tilde{x}} = \left[\mathbf{\tilde{k}}_{\text{com}}^{\intercal}, \mathbf{\tilde{q}}_b^{\intercal}, \mathbf{\tilde{q}}_j^{\intercal}\right]^{\intercal}$ where $\mathbf{\tilde{q}}_b$ can be intuitively described as the concatenated base linear and angular velocity unit vectors required to transform the body from its initial pose to its current pose. This transformation can be described with pose exponential maps.





After some derivations \cite{galileo}, we have 
\begin{equation}
    \mathbf{F}_{\text{exp}}(\mathbf{x}(t), \mathbf{\tilde{x}}(t)) = \begin{bmatrix}
                        &\mathbf{k}_{\text{com}}(t) + \mathbf{\tilde{k}}_{\text{com}}(t)\\
                        &\mathbf{p}(t) + \mathbf{q}_{bo}(t)\mathbf{R}\mathbf{v}(t)\\
                        &\mathbf{q}_{bo}(t)\otimes \text{exp}_{q_{bo}}(\mathbf{q}_{bo,\omega}(t))\\
                        &\mathbf{q}_j + \mathbf{\tilde{q}}_j
                    \end{bmatrix}
\end{equation}


%% file: Results.tex
\section{Results}\label{results}


To showcase the feasibility of our method, we developed Galileo \cite{galileo}, a powerful C++ MPC framework for legged robots, and tested our approach in simulation (Gazebo) for the Unitree Go1 quadruped. Our work results in 50Hz MPC trajectories, enabling real-time locomotion. For the whole-body controller and perception, we use the framework introduced in \cite{grandia2022perceptivelocomotionnonlinearmodel} to track our MPC trajectories and generate steppable region constraints, which allows us to perform stairclimbing behaviors as shown in Fig. \ref{fig:galileo-results}. Results can be viewed at \cite{galileo}.

%% file: Conclusions.tex
\section{Conclusions and Future Work}\label{conclusions}
We presented Galileo \cite{galileo}, a new C++ library for robot trajectory optimization, using a modified pseudospectral collocation method to optimize over state manifolds such as SE(3). In future work, we would like to pursue two distinct avenues: 1) Automating the generation of initial guesses to improve convergence speed, and 2) Presenting a comparison of Galileo with state-of-the-art locomotion frameworks such as \cite{Farshidian2017OCS2} and \cite{Crocoddyl}.